\pdfoutput=1

\documentclass[11pt]{article}

\usepackage[]{EMNLP2023}

\usepackage{times}
\usepackage{latexsym}
\usepackage{multirow}
\usepackage{graphicx}
\usepackage{amsmath}
\usepackage[T1]{fontenc}
\usepackage{tabularx}
\usepackage[utf8]{inputenc}

\usepackage{microtype}

\usepackage{inconsolata}

\title{CommunityKG-RAG: Leveraging Community Structures in Knowledge Graphs for Advanced Retrieval-Augmented Generation in Fact-Checking}

\author{Rong-Ching Chang \\
  Department of Computer Science \\
  University of California, Davis \\
  Davis, CA \\
  \texttt{rocchang@ucdavis.edu} \\\And
  Jiawei Zhang \\
  Department of Computer Science \\
  University of California, Davis \\
  Davis, CA \\
  \texttt{jiawei@ifmlab.org} \\}

\begin{document}
\maketitle
\begin{abstract}

Despite advancements in Large Language Models (LLMs) and Retrieval-Augmented Generation (RAG) systems, their effectiveness is often hindered by a lack of integration with entity relationships and community structures, limiting their ability to provide contextually rich and accurate information retrieval for fact-checking. We introduce CommunityKG-RAG (Community Knowledge Graph-Retrieval Augmented Generation), a novel zero-shot framework that integrates community structures within Knowledge Graphs (KGs) with RAG systems to enhance the fact-checking process. 
Capable of adapting to new domains and queries without additional training, CommunityKG-RAG utilizes the multi-hop nature of community structures within KGs to significantly improve the accuracy and relevance of information retrieval. Our experimental results demonstrate that CommunityKG-RAG outperforms traditional methods, representing a significant advancement in fact-checking by offering a robust, scalable, and efficient solution.
\end{abstract}

\section{Introduction}


The occurrence of misinformation and the imperative of fact-checking are pivotal elements within the digital information ecosystem, profoundly affecting public discourse and shaping societal decisions worldwide. Concurrently, the advent of Large Language Models (LLMs) has unveiled remarkable capabilities in comprehending and producing human languages, presenting a promising avenue for bolstering fact-checking endeavors.
Prior research \cite{buchholz2023assessing,li2023self,caramancion2023harnessing,hoes2023leveraging,huang2023harnessing}  has delved into directly prompting LLM models to identify false information. 
However, while LLMs can be instrumental in combating misinformation, their practical application still exposes two critical limitations. 
Firstly, these models are constrained by the cut-off date of their training data. 
Secondly, this issue is compounded by the tendency of LLMs to generate incorrect information or ``hallucinations” \cite{huang2023survey} which could jeopardize the accuracy of claim verification in fact-checking tasks.

In response to these challenges, Retrieval-Augmented Generation (RAG) has emerged as a promising approach. By integrating the generative capabilities of LLMs with external data retrieval, RAG significantly enhances the accuracy and relevance of the responses. 
%
For instance, \citet{Liao_2023} leverages RAG by employing both the dot product and the BERT-based sequence tagging model to identify key evidences.
 \citet{soleimani2019bert} uses the BERT model to retrieve and validate claims. 

While RAG significantly advances the capabilities of LLMs, it, too, faces unique challenges. 
Firstly, language models suffer from utilizing contexts in long texts.  
When crucial information is located in the middle, it is less likely to be effectively utilized by language models \cite{liu2023lost}. 
Secondly, when contexts are laden with noise or contradictory information, RAG's performance can be adversely underscored \cite{barnett2024seven}. 
Thirdly, the retrieval process plays a crucial role. Often, even if the answer to a query is present in the document corpus, it may not rank highly enough to be returned to the user \cite{barnett2024seven}.
Further expanding on the challenges in RAG systems, 
knowledge retrieved by these systems does not always contribute positively  \cite{wang2023self} and can sometimes detrimentally impact the original responses generated by the LLMs.
%

Acknowledging the challenges inherent in RAG systems, Knowledge Graphs (KGs) offer a structured, semantically rich framework that has a long-standing history of enhancing fact-checking efforts.
KGs play a crucial role in encapsulating and organizing complex information through their inherent structure which is comprised of triples. Each triple, consisting of a subject, predicate, and object — alternatively framed as a head entity, a relation, and a tail entity \textit{i.e.}, (subject entity, relationship, object entity) — constitutes the core component of a KG, enabling it to represent structural facts and support symbolic reasoning effectively.

KGs represent data in a way that captures information about not just the entities but also the complex relationships between them. This semantic web of information allows for a deeper understanding of context, which is essential for verifying facts.
Furthermore, KGs facilitate the exploration of multi-hop information pathways, allowing for the elucidation of intricate and indirect relationships critical for comprehensive fact verification. 
Prior work has shown promising results utilizing KGs \cite{hu2023givememoredetails,liu-etal-2020-fine,ma-etal-2023-kapalm}. 
 %
However, concurrently integrating both the structured knowledge graphs with unstructured text as inputs to LLMs is not a trivial enterprise. 
Prior work has tried directly including triples as input to LLMs \cite{baek2023knowledge,sequeda2023benchmark}. Yet LLMs are not trained for leveraging triples, and this approach does not leverage the community and entity relationship. 
Other approaches \cite{sun2021ernie,liu2020k,yasunaga2022deep,sun2020colake,zhang2022greaselm,kang2023knowledge} require training customized models or joint embeddings that are computationally expensive.

In light of the distinct advantages of KGs and the capabilities of RAG systems and LLMs, the absence of research on their combined application for fact-checking is notable. Although such integration —— melding KGs' structured, semantic insights with RAG's dynamic retrieval and LLMs' language comprehension —— holds significant promise for advancing fact-checking technologies, the specific impact of this synergistic approach remains largely unexplored.

\begin{figure*}
    \centering
    \includegraphics[width=0.9\linewidth]{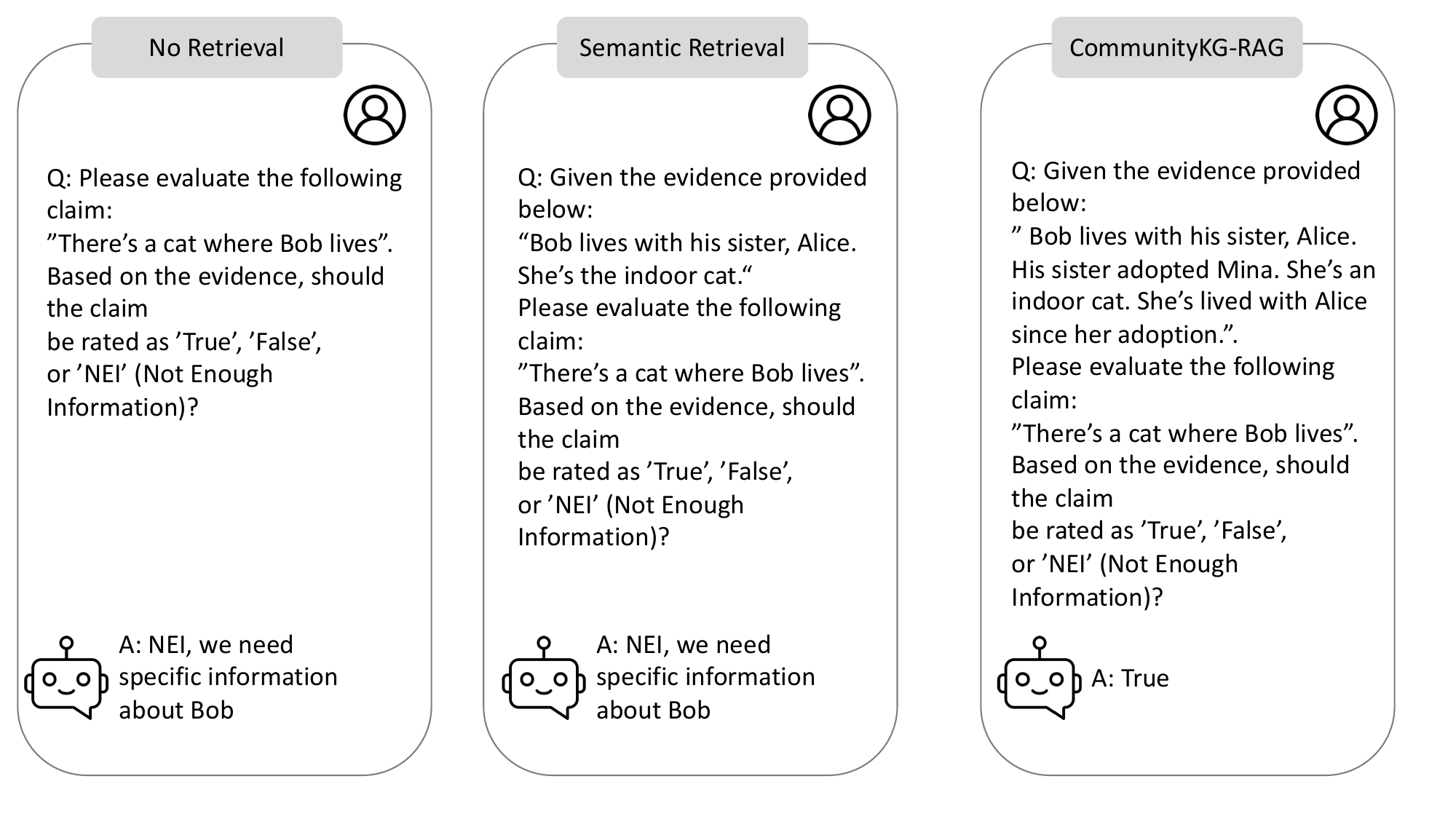}
    \caption{Comparison between no retrieval, semantic retrieval, and CommunityKG-RAG. The no retrieval and semantic retrieval fail to provide sufficient context, while our proposed method, CommunityKG-RAG, is able to by leveraging multi-hop knowledge graph information in the retrieval process enhancing accuracy and relevance.}
    \label{fig:demo}
\end{figure*}

To bridge the existing research gap, we introduce a pioneering framework: \textbf{CommunityKG-RAG (Community Knowledge Graph-Retrieval Augmented Generation)}. This innovative approach synergizes Knowledge Graphs with Retrieval-Augmented Generation and Large Language Models to enhance fact-checking capabilities. By leveraging and preserving the intricate entity relationships and community structures within KGs, our framework provides a contextually enriched and semantically aware retrieval mechanism that significantly improves the accuracy and relevance of generated responses. Specifically, we construct a comprehensive KG from fact-checking articles, employ the Louvain algorithm for community detection, and assign embeddings derived from word embeddings to each node. This approach ensures that the identified communities are both structurally coherent within the KG and highly pertinent to the fact-checking task. By harnessing this integrated framework, we offer a robust, scalable, and efficient solution to contemporary fact-checking challenges. 
An example of this integration and its impact on retrieval accuracy is illustrated in Figure \ref{fig:demo}.

Our contributions are threefold:

\begin{enumerate}
    \item \textbf{Utilization of Both Structured and Unstructured Data with Superior Knowledge Graph Integration:} By combining the structured data of Knowledge Graphs with the unstructured data handled by LLMs, we achieve a more comprehensive and context-aware fact-checking system. We demonstrate that converting knowledge graphs back to sentences within our framework is superior to methods that use triples as context. This approach enhances the comprehensibility and relevance of the retrieved information, as demonstrated by the significant increase in accuracy.
    
    \item \textbf{Context-Aware Retrieval and Multi-hop Utilization:} By leveraging community structures and multi-hop paths within KGs, the framework delivers more precise and relevant information retrieval, enhancing the overall effectiveness of the fact-checking process. We are the first work to propose utilizing and combining multi-hop in KGs with RAG systems. 
    \item \textbf{Scalability and Efficiency:} The framework operates in a zero-shot manner, requiring no additional training or fine-tuning, which ensures high scalability and adaptability to various LLMs. Additionally, the knowledge graph and community detection processes only need to be performed once, allowing for repeated reuse or rapid updates.

\end{enumerate}

\section{Related Work}




\textbf{KGs in LLM inputs}


Recent research has explored the integration of KGs with LLMs, where triples are directly fed into LLMs as input \cite{baek2023knowledge,sequeda2023benchmark}.
However, this approach has its limitations, particularly in its assumption that LLMs can effectively process and utilize triples despite their primary training focus on sequential data processing. 
This could result in an underutilization of KG's structural information, such as subgraph structure, community structure, and relationship patterns across entities and relations of Knowledge Graphs. 
%
Addressing this, our proposed method leverages community detection results as indices for text retrieval, thus harnessing the subgraph and entity relationship structures inherent in KGs more effectively than in previous work.

Other approaches to integrating knowledge graphs with language models include joint embedding training or the customization of model architectures. 
This can be done by representing triplets as a sequence of tokens and concatenating them with text embedding in the pre-training stage \cite{sun2021ernie,liu2020k}. For instance, 
\citet{yasunaga2022deep} propose a cross-modal model to fuse text and KG to jointly pre-train the model.
\citet{sun2020colake} present a word-knowledge graph that unifies words and knowledge.  
\citet{zhang2022greaselm} fuses representations from pre-trained language models and graph neural networks over multiple layers.
Models that require additional training are computationally expensive and cumbersome. 
\citet{kang2023knowledge} retrieves a relevant subgraph composed of triples by utilizing GNN for triple embedding.
In contrast, our method does not necessitate additional training, offering a more efficient and adaptable solution for integrating KGs with LLMs.

\section{Problem Statement}

The goal of fact-checking task formulation is to locate the top $n$ most relevant sentence, in order to classify a given claim as either \textit{refuted}, \textit{supported}, or \textit{not enough information} as the labels by a large language model.
Let $ P$ represent a corpus of fact-checking articles and $ {C}$ a set of claims.
Each claim $ c \in C$ is associated with a ground-truth label $ y$. 
There exists a set of top $ k$ most relevant 
 sentences $ P_c = {p}_{i}^k$ from the fact-checking articles $ P$ for each claim $ c$.  
The task is formulated as optimizing the prediction $ \hat{y} = f(C,P_c)$, where $ f$ is a large language model to evaluate the truthfulness of claims based on the evidence provided.

\begin{figure*}
    \centering
    \includegraphics[width=\linewidth]{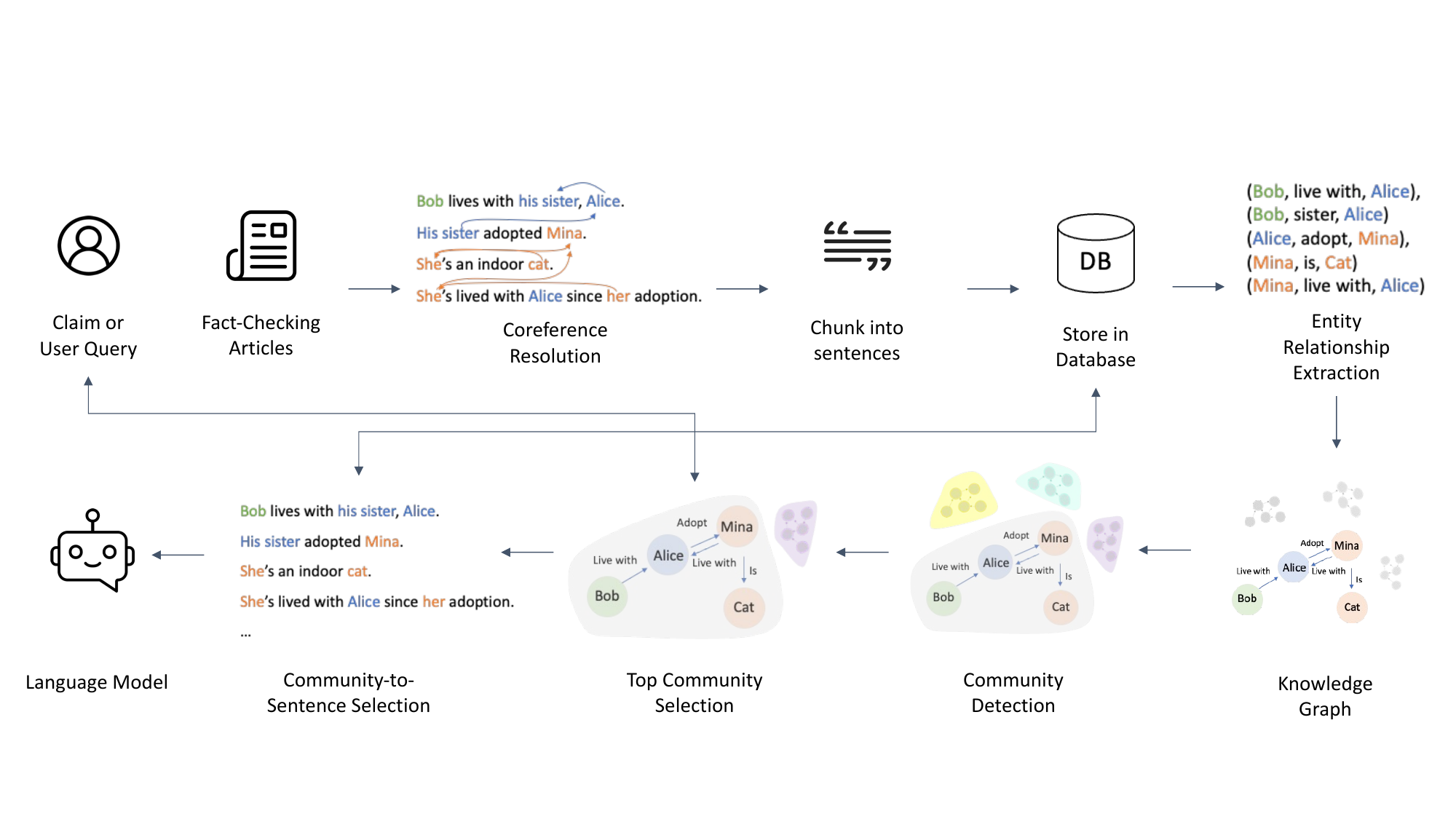}
    \caption{Workflow of CommunityKG-RAG}
    \label{fig:archi}
\end{figure*}

\section{CommunityKG-RAG}

In this section, we detail our novel framework CommunityKG-RAG for integrating KGs with RAG systems and LLMs to enhance fact-checking capabilities. We show an overview in Figure \ref{fig:archi}.
Our approach leverages the structural advantages of KGs to provide a contextually enriched, semantically aware information retrieval mechanism, which is subsequently used to inform the generation process of LLMs.

\subsection{Knowledge Graph Construction}

We begin by constructing a KG from a corpus of fact-checking articles. The construction process involves the following three steps:

\subsubsection{Coreference Resolution}
Coreference resolution is a preprocessing step to enhance the semantic coherence of the input data prior to knowledge graph construction. This process aims to identify and cluster mentions of entities and pronouns that refer to the same real-world entities across the corpus, thereby resolving ambiguities in entity references.

We employ a state-of-the-art coreference resolution model by \citet{lee2018higherorder}, leveraging a deep learning approach based on SpanBERT \cite{10.1162/tacl_a_00300}, which has been pre-trained on a large corpus to capture a wide range of syntactic and semantic information.

\subsubsection{Graph Construction}

CommunityKG-RAG leverages the relationship extraction model, REBEL, proposed by \citet{cabot2021rebel} to discern entity relationships within the corpus. This process is formalized as follows:

Given the corpus $ P$, we extract a set of entities, denoted as \(E = \{e_1, e_2, ..., e_n\}\). 
We construct the entity graph \(G = (E, R)\), where \(R\) comprises the set of relationships between entities. In this graph, entities (\(E\)) are represented as nodes, and relationships (\(R\)) are depicted as edges that link these nodes. This graph represents the intricate network of connections among entities derived from the corpus, forming the foundation of the KG.

This structured approach facilitates a comprehensive representation of the factual relationships within articles, thereby enabling advanced analysis and application in fact-checking and misinformation identification tasks.

\subsubsection{Node Feature Embedding}\label{sec:node_feature} For each node in the KG, we assign it with word embeddings derived from a pre-trained BERT model \cite{BERT}. This embedding serves as the node feature vector, encapsulating the semantic information of the entity.

\subsection{Community Detection} 

To leverage the community structures inherent within the Knowledge Graph (KG) for enhanced fact retrieval, we employ the Louvain algorithm \cite{Blondel_2008} as a foundational tool. This algorithm is instrumental in detecting and delineating communities within the graph \( G \), by focusing on the optimization of modularity. Modularity is a scalar value between \(-1\) and \(1\) that measures the density of links inside communities compared to links between communities. The algorithm initially treats each node as its own community and iteratively merges communities to maximize the gain in modularity. This optimization continues until no further improvement in modularity is possible, resulting in a partition of the graph into distinct communities. 

From graph \( G \), we extract a set of communities denoted by \( M \), where each community \(m \in M \) represents a cluster of nodes more interconnected among themselves than with the rest of the graph. This structured approach allows us to focus our retrieval efforts on specific segments of the KG that are more likely to contain relevant and contextually rich information for fact-checking tasks.

\subsection{Community Retrieval}
Each community $ m$ is considered as a subgraph $ G_m = (E_m, R_m)$ comprising a subset of entity nodes $ E_m$ and their relationships $R_m$. 
The embedding representation of each community denoted as \( \varphi(m) \) is derived by averaging the BERT embeddings of the nodes within $ E_m$:

\[
\varphi(m) = \frac{1}{|E_m|} \sum_{i \in E_m} \text{BERT}(e_i)
\]

where \( |E_m| \) is the number of nodes in a community \( m \) and \( e_i \) represents the word embedding of node \( i \) derived from BERT model \cite{li2023blip} as described in the section \ref{sec:node_feature}. 
This approach aggregates the collective semantic attributes of the community, encapsulating a comprehensive semantic representation.

To convert claims into embeddings for similarity comparisons, we utilize the BERT-base Sentence Transformer model, Sentence-BERT \cite{reimers-2019-sentence-bert}. Sentence-BERT is specifically optimized for generating high-quality sentence embeddings, making it ideally suited for comparing the semantic similarities between claims and community descriptions.

The relevance score $ r(c,m)$ between claim $ c$ and community $ m$ is calculated 
as the dot product between their embeddings:
    
\[
r(c,m) = \varphi(c)^T\varphi(m)
\]

\subsection{Top Community Selection}

To efficiently prioritize communities for deeper analysis, the top \( \delta \) percent of communities, ranked by their relevance scores \( r(c, m) \), are selected. The selection threshold \( N \) is determined as follows:
$ N = \left\lceil \frac{\delta}{100} \times |M| \right\rceil $, where \( |M| \) represents the total number of communities. 
Consequently, the subset of most relevant communities \( M^*_c \) to claim $ c$ is defined as:

\[
M^*_c = \{ m \in M : \text{rank}(r(c, m)) \leq N \}
\]

This selection criterion ensures that our analysis is concentrated on the communities most likely to contain relevant and substantive information pertinent to the verification of a claim \( c \), thus facilitating efficient and focused fact-checking.

\subsection{Top Community-to-Sentence Selection}

To identify the most pertinent sentences, a relevance score \( r(M^*_c,p) \) is computed for each sentence \( p \) within the top communities \( M^*_c \). Sentences are then ranked by relevance, and the top \( \lambda \) percent are selected, resulting in a subset \( P^*_c \) of the most relevant sentences. 

This structured approach allows for systematic filtering and selection of significant information, a process which is crucial for robust and focused fact-checking.
We use \( \text{CommunityKG-RAG}^{\delta}_{\lambda} \) to represent the synergistic application of two distinct filters: the top \( \delta \) percent for community relevance and the top \( \lambda \) percent for sentence significance within the context of validating community-to-sentence relevance. This refined designation underscores a strategic methodological synthesis aimed at optimizing the fact-checking process by methodically concentrating on the most pivotal communities and their essential corresponding sentences.

\section{Experimental Details}

\subsection{Datasets}

\textbf{MOCHEG} This multimodal fack-checking dataset \cite{menglong2022end} consists of 15,601 claims annotated with a truthfulness label collected from PolitiFact and Snopes, two popular websites for fact-checking articles. The articles and results of claim verification were produced by journalists manually. The truthfulness is labeled into three categories: supported, refuted, and NEI (not enough information). More details are included in the Appendix \ref{sup:data}.

\subsection{Baselines}
\textbf{No Retrieval}
This is a naive baseline where answers are generated from the language model through prompts without context or retrieval.

\textbf{Semantic Retrieval} Following \citet{nie2019combining}, we extract context based on semantic similarity. Specifically, we use cosine similarity in embeddings between the prompt and the context. BERT \cite{devlin2018bert} is used to produce the embedding. 

\textbf{Knowledge-Augmented language model PromptING (KAPING)} We implement KAPING proposed by \citet{baek2023knowledge}. The KAPING is a zero-shot RAG framework that proposes basing retrieval on sentence similarity between the input text and triples. 
The output prompt of the KAPING framework includes the original text prompt with triples as the context. Specifically, the triples are in the format of $ (subject entity, relationship, object entity)$. We equip KAPING with the same set of articles for retrieval.

\subsection{Implementation Details}
We conducted our experiments using the LLaMa2 7 billion model as our primary Large Language Model \cite{touvron2023llama}. The LLaMa2 models are open-source and widely accessible. We chose these models because they were trained on trillions of tokens, including publicly available datasets like Wikipedia, and demonstrated state-of-the-art results at the time when the texts were published. This capability enabled a thorough evaluation of our method's zero-shot performance when applied to previously unseen corpora.

The availability of these models in multiple sizes enabled a comparative analysis of our proposed framework, assessing how model scale impacts performance. Furthermore, since Wikipedia was integral to their training datasets, we were able to explore the efficacy of our approach on corpora familiar to the models. The utility of this retrieval approach has been substantiated in prior research \cite{khandelwal2020generalization}.

To quantitatively assess the LLMs, we measured their performance in verifying claims using accuracy as our metric. More details of the LLMs and the corresponding prompt are included in Appendices \ref{app:pmp} and \ref{app:llp}.

We use $\text{CommunityKG-RAG}^{25}_{100}$ as the baseline. In other words, we use the top $ \delta = 25$ percent of the most relevant communities and $\lambda = 100$ percent of the sentences that the community maps to as the context.

\section{Results}

\begin{table}[]
    \centering
    \begin{tabular}{cc}
    \hline
        Model &  LLaMa2 7B\\
        \hline
         No Retrieval & 39.79\%\\
         Semantic Retrieval  & 43.84 \%\\
         KAPING & 39.41 \% \\
         \hline
         $\text{CommunityKG-RAG}^{25}_{100}$ & \textbf{56.24\%}   \\
         \hline
    \end{tabular}
\caption{Comparison of claim verification accuracy for various retrieval methods: No Retrieval, Semantic Retrieval, KAPING, and our approach, $\text{CommunityKG-RAG}^{25}_{100}$, which selects the top 25 percent of relevant communities and uses 100 percent of their mapped sentences as context. }
    \label{tab:accuracy}
\end{table}

\subsection{Main Results}
Overall, our proposed method, \( \text{CommunityKG-RAG}^{25}_{100} \), not only achieves the best results but also surpasses all baselines, as detailed in Table \ref{tab:accuracy}. The No Retrieval baseline recorded an accuracy of 39.79 percent. Employing the Semantic Retrieval strategy yielded an improvement, elevating accuracy to 43.84 percent. This increase underscores the advantages of integrating semantic context, thereby enhancing the proficiency of the language model in claim verification.

Conversely, the KAPING method did not enhance performance, registering a slight decline in accuracy to 39.41 percent. This outcome indicates that a language model such as LLaMa2 may struggle with retrieval contexts formatted as triples (\textit{i.e.}, (subject entity, relationship, object entity) ). Such structuring appears to impede the model's capacity to effectively utilize information. This is likely due to its foundational training on sequential word prediction rather than on processing structured data like triples.

However, the performance of our approach, \( \text{CommunityKG-RAG}^{25}_{100} \), was markedly superior, achieving an accuracy of 56.24 percent. This significant increase not only confirms the effectiveness of integrating community-derived knowledge into the retrieval process but also demonstrates substantial gains over conventional retrieval methods. These results validate the substantial impact that tailored, community-focused retrieval mechanisms can have on the operational effectiveness of language models in complex verification scenarios. This marked improvement reiterates the critical role of precise, context-aware retrieval strategies in augmenting the functional capabilities of language models.

\subsection{Ablation}

We conducted a series of ablation studies to understand the significance of various factors within the \text{CommunityKG-RAG} framework. Specifically, these ablation studies are designed to evaluate the impact of different backbone language models, the selection of top communities, and the extent of community-to-sentence selection.

\subsubsection{Performance With Different Backbone Models}
To demonstrate the robustness and adaptability of the proposed \text{CommunityKG-RAG} framework, we conducted an ablation study to assess how different backbone language models affect the performance on the MOCHEG fact-checking dataset. 
Considering the computational costs, which increase with the number of communities and community-to-sentences selection using the community (Appendix \ref{app:community}), we conduct this ablation with $\text{CommunityKG-RAG}^{25}_{25}$. We selected the top $ \delta = 25 $ percent of the most relevant communities and the top $\lambda = 25$ percent of the sentences mapped to these communities to serve as the contextual input.

In this analysis, we compared the performance of two different backbone models: LLaMa2 7B and LLaMa3 8B. Table \ref{tab:backbone_models_performance} illustrates the outcomes, showing that \text{CommunityKG-RAG} significantly enhances performance across both models. Specifically, when employing the \text{CommunityKG-RAG} framework, there is a notable improvement of 6.18 percentage points with LLaMa2 7B and an increase of 3.21 percentage points with LLaMa3 8B compared to the no retrieval baseline. 
However, we observed that the LLaMa3 8B showed a lesser improvement and accuracy over the no retrieval baseline than the 7B model despite its larger size. This may be attributed to the 8B model's capability to explore various facets of a given issue more comprehensively, which, while generally beneficial, might lead to a less precise matching in scenarios demanding exact binary evaluations, such as our fact-checking tasks. This characteristic could also contribute to the slightly lower improvement observed with the 8B model.

These results underscore the effectiveness of our framework in leveraging structured community knowledge, thereby improving the accuracy of fact-checking across diverse language model architectures.

\begin{table}[ht]
    \centering
    \begin{tabular}{|c|c|c|}
        \hline
        \textbf{Model} & \textbf{LLaMa2} & \textbf{LLaMa3} \\
         & \textbf{7B} & \textbf{8B} \\
        \hline
        \textbf{No Retrieval} & 39.79\% & 26.03\% \\
        \textbf{$\text{CommunityKG-RAG}^{25}_{25}$} & 45.97\% & 29.24\% \\
        \hline
    \end{tabular}
    \caption{Performance comparison of no retrieval and CommunityKG-RAG with $\delta = 25$ and $\lambda = 25$ settings across different backbone models, LLaMa2 7B and LLaMa3 8B.}
    \label{tab:backbone_models_performance}
\end{table}

\subsubsection{Influence of Community-to-Sentence Selection}

This section examines the influence of varying community-to-sentence selection thresholds within a consistently held community threshold of 25 percent on the performance of the CommunityKG-RAG framework using the LLaMa2 7B model. Community-to-sentence selection thresholds were adjusted to 25 percent, 50 percent, 75 percent, and 100 percent to identify the optimal level for enhancing fact-checking performance.

\begin{table}[ht]
\centering
\begin{tabular}{cc}
\hline
Model & LLaMa2 7B \\
\hline
$\text{CommunityKG-RAG}^{25}_{25}$ & 45.97\% \\
$\text{CommunityKG-RAG}^{25}_{50}$ & 27.83\% \\
$\text{CommunityKG-RAG}^{25}_{75}$ & 41.93\% \\
$\text{CommunityKG-RAG}^{25}_{100}$ & 56.24\% \\
\hline
\end{tabular}
\caption{Performance variations of the LLaMa2 7B model under the CommunityKG-RAG framework with consistent community threshold (top 25 percent) and variable community-to-sentence selection. }
\label{tab:abla_sent_accuracy}
\end{table}

The results presented in Table \ref{tab:abla_sent_accuracy} demonstrate variable model performance as community-to-sentence selection thresholds change. Initially, the performance slightly decreases to 27.83 percent when the inclusion rate of sentences is increased from 25 percent to 50 percent. This might indicate that the top 25 percent of sentences contain the most crucial information for verifying the claim, and including additional sentences up to 50 percent introduces noise or less relevant data that temporarily hinder the model’s accuracy. However, as the inclusion rate continues to increase to 75 percent and then to 100 percent, the performance improves, ultimately achieving the highest accuracy at a full 100 percent inclusion rate. This suggests that beyond the 50 percent threshold, the additional sentences contribute positively, possibly by providing necessary context that supports more accurate fact-checking.

This pattern highlights the critical role of extensive contextual engagement in the CommunityKG-RAG framework, demonstrating that access to a wider array of sentences associated with a carefully selected group of communities markedly improves the model's effectiveness in accurately identifying truth and falsehood. These results underscore the nuanced balance needed in selection strategies to provide adequate context for accurate analysis without inundating the model with extraneous data.

\subsubsection{Combined Effects of Top Community and Community-to-Sentence Selection}
To further explore the efficacy of the CommunityKG-RAG framework, we conducted an analysis to understand the impact of varying the top community and community-to-sentence selection thresholds on the performance of the model. We adjusted the thresholds of both $\delta$ and $\lambda$ to 25 percent, 50 percent, 75 percent, and 100 percent to examine how the extent of considered context in both community and community-to-sentence selection affect the fact-checking capabilities of the CommunityKG-RAG framework. We show the knowledge graph community statistics in Appendix \ref{app:community}. 

The results, as shown in Table \ref{tab:abla_commu_accuracy},  reveal interesting trends. Initially, the increase of thresholds from 25 percent to 75 percent led to a slight decrease in performance, suggesting that adding more communities and sentences might introduce noise or less relevant information, thus compromising the model's effectiveness. However, a significant improvement is observed when the thresholds are expanded to 100 percent. This enhancement at the highest threshold suggests that the model benefits from a more comprehensive view of the available data, possibly capturing essential contextual nuances that are otherwise missed at lower thresholds. This pattern aligns with observations from previous ablation studies concerning community-to-sentence selection.

Interestingly, when comparing the effects of top community selection, an increase in the number of top communities results in improved accuracy while holding community-to-sentence selection constant. This observation emerges from comparing $\text{CommunityKG-RAG}^{25}_{50}$ versus $\text{CommunityKG-RAG}^{50}_{50}$, and $\text{CommunityKG-RAG}^{25}_{75}$ to $\text{CommunityKG-RAG}^{75}_{75}$.

However, increasing both the community selection and community-to-sentence selection to 100 percent does not yield further improvements. As illustrated by the comparison between $\text{CommunityKG-RAG}^{25}_{100}$ and $\text{CommunityKG-RAG}^{100}_{100}$, this finding implies that a targeted selection of highly relevant communities, along with a comprehensive examination of their associated sentences, strikes an ideal balance. It enables the model to access detailed and pertinent information effectively without being overwhelmed by extraneous data. This method provides a nuanced approach to information retrieval that maximizes accuracy while avoiding information overload.

\begin{table}[]
    \centering
    \begin{tabular}{cc}
    \hline
        Model &  LLaMa2 7B\\
        \hline

         $\text{CommunityKG-RAG}^{25}_{25}$ & 45.97\%   \\
          $\text{CommunityKG-RAG}^{50}_{50}$ & 43.64\%   \\
          $\text{CommunityKG-RAG}^{75}_{75}$ & 43.60\%   \\
          $\text{CommunityKG-RAG}^{100}_{100}$ & 54.62\%   \\
         \hline
    \end{tabular}
\caption{Performance metrics of the LLaMa2 7B model within the CommunityKG-RAG framework across varied thresholds of top community and community-to-sentence selection. The table details the model's accuracy percentages at incremental selection thresholds of 25, 50, 75, and 100 percent for both community and community-to-sentence selection, illustrating how varying levels of context inclusion impact the model’s performance.}
    \label{tab:abla_commu_accuracy}
\end{table}

\section{Conclusion}

We have introduced CommunityKG-RAG, a novel framework that integrates Knowledge Graphs with Retrieval-Augmented Generation and Large Language Models to enhance fact-checking. This approach leverages the structured data of KGs and the generative capabilities of LLMs, significantly improving the accuracy and relevance of responses.

CommunityKG-RAG effectively addresses key challenges such as outdated information and hallucinations by utilizing multi-hop community structures for refined and accurate retrieval within KGs. This integration enables more precise and contextually rich information retrieval, crucial for effective fact-checking. Our framework achieves superior performance without requiring any fine-tuning or additional training, demonstrating its robustness and efficiency. As the first framework to combine multi-hop community information in KGs with RAG systems, CommunityKG-RAG represents a significant advancement and promising direction for future work.

\section{Limitations}

Despite the notable success of the CommunityKG-RAG framework in enhancing claim verification accuracy, several limitations highlight areas for future research and improvement:

\subsection{Computational Demands}
The CommunityKG-RAG framework places substantial demands on computational resources compared to no retrieval or semantic retrieval. However, communities can be pre-computed and reused, making the operational phase more lightweight and dynamic. This capability enhances the model's responsiveness to new data and trends. Further, our method has demonstrated significant accuracy improvements despite the computational demands, and, besides, our proposed method is more lightweight than methods that require training or fine-tuning a language model.

\subsection{Dependency on Entity Recognition Quality}
Our proposed method's effectiveness heavily relies on the quality of entity recognition. 
There are prior works \cite{edge2024local} that rely on utilizing language models to conduct entity recognition. This could potentially introduce hallucinations. To avoid such risk, we use REBEL, a seq2seq model based on Wikipedia data. If the framework is applied to text that is significantly different from Wikipedia text, it might hinder performance. In such cases, utilizing an entity recognition method tailored to the specific domain could be beneficial. However, as shown in the Appendix \ref{app:community}, our approach incorporates a comprehensive dataset with up to 48,630 nodes and 202,455 edges, which ensures a robust and extensive knowledge base. This comprehensive coverage helps mitigate potential quality issues, enhancing the reliability of the entity recognition process.

These limitations, alongside the outlined implementation advantages, underscore the need for ongoing refinement and testing of the CommunityKG-RAG framework to optimize its practicality and effectiveness in real-world scenarios. The ability to pre-compute communities ensures that the method remains operationally lightweight and scalable, an essential factor for broad application. Additionally, future work can consider extending this method framework into multimodality, integrating multimodal graphs or tabular data. Such extensions could further enhance the model's capabilities and applicability in more complex and varied data environments, opening new avenues for research and practical implementation.

\bibliography{ref}
\bibliographystyle{acl_natbib}

\appendix

\section{Details of Datasets}\label{sup:data}

The dataset was partitioned into training and testing subsets, with the training set employed for constructing the knowledge graph and verifying claim accuracy. Comprising 18,553 unique claims, each is linked to a corresponding fact-checking article and label.

The target variable, "truthfulness," is classified into three categories: "Supported," "Refuted," and "Not Enough Information" (NEI). The label distribution includes 7,137 "Refuted," 6,928 "Supported," and 4,488 "NEI."

Label assignment for "Supported," "Refuted," and "NEI" was performed following a meticulous cleaning process carried out by the authors of \citet{menglong2022end}. This process was conducted as the original labels derived from the fact-checking articles were marred by noise and inconsistency. Initially, the labels encompassed a broad spectrum of classifications, including "False," "Mostly False," and "Half True," totaling up to 75 different labels. This refinement was crucial as the original articles did not explicitly categorize claims into "Supported," "Refuted," or "NEI." This ambiguity could potentially impair the retrieval capabilities of large language models (LLMs). To mitigate this, we simplified the labels by mapping "Supported" to "True" and "Refuted" to "False" during the prompting and preprocessing phases.

\section{Prompt}\label{app:pmp}
The prompt used for all RAG systems is the following: 

\begin{minipage}{\columnwidth}
\texttt{"Given the evidence provided below: \{formatted\_evidence\}.\\
Please evaluate the following claim: \{claim\}.\\
Based on the evidence, should the claim be rated as 'True', 'False',\\
or 'NEI' (Not Enough Information)?"}
\end{minipage}

The prompt used for all baseline zero shot setups is the following: 

\begin{minipage}{\columnwidth}
\texttt{"Please evaluate the following claim: \{claim\}.\\
Based on the evidence, should the claim be rated as 'True', 'False',\\
or 'NEI' (Not Enough Information)?"}
\end{minipage}

\section{Language Model Parameters}\label{app:llp}

In our experiments, we utilized the meta-llama/Llama-2-7b-chat-hf model from Hugging Face's model hub. Our generation pipeline was configured to produce coherent and non-repetitive text. Key settings included a temperature of 0.3 to encourage predictability, a repetition penalty of 1.1 to avoid redundant content, and a limit of 200 new tokens per output to maintain focus. Custom stopping criteria were implemented to end text generation at specific tokens, ensuring outputs remained within the scope of our conversational framework. 

\section{Computing Infrastructure}
All computational experiments were conducted on a server configured with two NVIDIA RTX A6000 GPUs, each with 48 GB of GDDR6 memory, and two AMD EPYC 7513 32-core processors. The system also included 512 GB of DDR4 ECC RAM and a 960 GB Samsung PM983 NVMe SSD for storage.

\section{Community Statistics}\label{app:community}
We provide the knowledge graph community statistics with various top \( \delta \) percent communities in Table \ref{tab:community_data}. These statistics demonstrate the multi-hop nature of our knowledge graphs through the metrics of average shortest path length and diameter. The average shortest path length, ranging from 4.03 to 4.28 across different community percentages, indicates that on average, multiple hops are required to traverse between nodes. The diameter values, ranging from 13 to 17, suggest the presence of long paths within the graphs, further supporting the existence of multi-hop pathways. These metrics confirm that our CommunityKG-RAG framework effectively leverages multi-hop connections, crucial for retrieving contextually rich and relevant information in fact-checking tasks.

\begin{table}[h]
    \centering
    \begin{tabular}{|p{4cm}|p{4cm}|}
        \hline
        \textbf{Metric} & \textbf{Value} \\
        \hline
        &\textbf{Top 25 Percent}   \\
        \hline
        Number of Nodes & 20,092 \\
        Number of Edges & 60,770 \\
        Avg. Degree & 6.05 \\
        Avg. Communities per Claim & 2.05 \\
        Avg. Nodes per Claim & 5.62 \\
        Avg. Shortest Path Length & 4.28 \\
        Diameter & 17 \\
        \hline
        &\textbf{Top 50 Percent}   \\
        \hline
        Number of Nodes & 32,428 \\
        Number of Edges & 117,677 \\
        Avg. Degree & 7.26 \\
        Avg. Communities per Claim & 4.57 \\
        Avg. Nodes per Claim & 11.63 \\
        Avg. Shortest Path Length & 4.13 \\
        Diameter & 13 \\
        \hline
        &\textbf{Top 75 Percent}   \\
        \hline
        Number of Nodes & 40,669 \\
        Number of Edges & 159,703 \\
        Avg. Degree & 7.85 \\
        Avg. Communities per Claim & 6.85 \\
        Avg. Nodes per Claim & 16.60 \\
        Avg. Shortest Path Length & 4.07 \\
        Diameter & 14 \\
        \hline
        &\textbf{Top 100 Percent}   \\
        \hline
        Number of Nodes & 48,630 \\
        Number of Edges & 202,455 \\
        Avg. Degree & 8.33 \\
        Avg. Communities per Claim & 9.64 \\
        Avg. Nodes per Claim & 22.25 \\
        Avg. Shortest Path Length & 4.03 \\
        Diameter & 13 \\
        \hline
    \end{tabular}
    \caption{Community Statistics}
    \label{tab:community_data}
\end{table}

\end{document}